\documentclass[journal,twoside,web]{ieeecolor}
\usepackage{tmi}
\usepackage[nocompress]{cite}

\let\oldcite\cite
\renewcommand{\cite}[1]{\textcolor{blue}{\oldcite{#1}}}

\usepackage{amsmath,amssymb,amsfonts}
\usepackage{algorithmic}
\usepackage{graphicx}
\usepackage{textcomp}

\usepackage{url}
\makeatletter
\let\NAT@parse\undefined
\makeatother
\usepackage{hyperref}
\hypersetup{
    colorlinks=true,
    linkcolor=black,
    citecolor=black,
    urlcolor=black,
    pdfborder={0 0 0}
}
\usepackage{booktabs}
\usepackage{array}

\usepackage{bm}

\usepackage{threeparttable}

\usepackage{fancyhdr}

\def\BibTeX{{\rm B\kern-.05em{\sc i\kern-.025em b}\kern-.08em
    T\kern-.1667em\lower.7ex\hbox{E}\kern-.125emX}}
\begin{document}
\title{Oral Imaging for Malocclusion Issues Assessments: OMNI Dataset, Deep Learning Baselines and Benchmarking}

\author{Pujun Xue,
        Junyi Ge,
        Xiaotong Jiang,
        Siyang Song$^*$,
        Zijian Wu,
        Yupeng Huo,
        Weicheng Xie,
        Linlin Shen,
        Xiaoqin Zhou,
        Xiaofeng Liu$^*$,
        and Min Gu$^*$
\thanks{This work was supported in part by the Top Talent of Changzhou “The 14th Five-Year Plan” High-Level Health Talents Training Project under Grant 2022260, in part by the Young Project of Changzhou Health Commission under Grant QN202310, in part by the Soochow University Graduate Education Reform Achievement Award Cultivation Project  under Grant KY20231517, in part by the National Natural Science Foundation under Grant 62276090, and in part by the International (Regional) Cooperation Project of Jiangsu Province under Grant BZ2024061. \textit{(*Corresponding Author: Min Gu, Xiaofeng Liu and Siyang Song.)}}
\thanks{Pujun Xue, Xiaoqin Zhou, and Xiaofeng Liu are with the College of Artificial Intelligence and Automation, Hohai University, Changzhou 213000, China (e-mail: pjxue@hhu.edu.cn; 20001453@hhu.edu.cn; xfliu@hhu.edu.cn).}
\thanks{Junyi Ge and Min Gu are affiliated with the Department of Stomatology, The Third Affiliated Hospital of Soochow University, and The First People’s Hospital of Changzhou, Changzhou 213003, China (e-mail: 527605767@qq.com; gumin106@163.com).}
\thanks{Xiaotong Jiang is with the College of Information Science and Engineering, Hohai University, Changzhou 213000, China (e-mail: 1747521405@qq.com).}
\thanks{Siyang Song is with HBUG lab, Department of Computer Science, University of Exeter, UK (e-mail: s.song@exeter.ac.uk).}
\thanks{Zijian Wu is with Nanjing University of Science and Technology, Nanjing, Jiangsu, and is a visiting student at HBUG lab, University of Exeter, UK (e-mail: wuzijian@njust.edu.cn).}
\thanks{Yupeng Huo is with Affect AI, Hefei, Anhui, and are visiting students with HBUG lab, University of Exeter, UK (e-mail: yphuo27@gmail.com).}
\thanks{Weicheng Xie and Linlin Shen are with Shenzhen University (e-mail: wcxie@szu.edu.cn; llshen@szu.edu.cn).}
}

\maketitle
\thispagestyle{fancy}

\fancyfoot[C]{\footnotesize This work has been submitted to the IEEE for possible publication. Copyright may be transferred without notice, after which this version may no longer be accessible.}

\begin{abstract}
Malocclusion is a major challenge in orthodontics, and its complex presentation and diverse clinical manifestations make accurate localization and diagnosis particularly important. Currently, one of the major shortcomings facing the field of dental image analysis is the lack of large-scale, accurately labeled datasets dedicated to malocclusion issues, which limits the development of automated diagnostics in the field of dentistry and leads to a lack of diagnostic accuracy and efficiency in clinical practice. Therefore, in this study, we propose the Oral and Maxillofacial Natural Images (OMNI) dataset, a novel and comprehensive dental image dataset aimed at advancing the study of analyzing dental images for issues of malocclusion. Specifically, the dataset contains 4166 multi-view images with 384 participants in data collection and annotated by professional dentists. In addition, we performed a comprehensive validation of the created OMNI dataset, including three CNN-based methods, two Transformer-based methods, and one GNN-based method, and conducted automated diagnostic experiments for malocclusion issues. The experimental results show that the OMNI dataset can facilitate the automated diagnosis research of malocclusion issues and provide a new benchmark for the research in this field. Our OMNI dataset and baseline code are publicly available at \url{https://github.com/RoundFaceJ/OMNI}.
\end{abstract}


\hypersetup{
    colorlinks=true,
    linkcolor=blue,
    citecolor=blue,
    urlcolor=blue,
    pdfborder={0 0 0}
}

\section{Introduction}
\label{sec:introduction}

\IEEEPARstart{M}{alocclusion} \cite{zhang2006impact} is an abnormal dental condition where teeth fail to align correctly during occlusion, resulting in functional and aesthetic issues between the teeth, jaws, and soft tissues of the face. Traditional diagnosis relies on the dentist’s experience and visual observation, which are subject to factors \cite{hassan2007occlusion} like professional knowledge, diagnostic fatigue, and lighting, leading to inconsistent results and misdiagnosis \cite{albalawi2022trends}. In addition, many inexperienced physicians lack the ability to handle complex cases (e.g., severe jaw deformity, multiple misaligned teeth, etc.) \cite{kapila2011current}, which adds to the difficulty and uncertainty of diagnosis. These limitations have prevent the development of automated malocclusion issues diagnostic methods, particularly deep learning (DL)-based methods. 

Nowadays, DL-based methods have been widely developed for various health-related assessment \cite{miotto2018deep,song2020spectral,shamshirband2021review} as they provide objective and repeatable diagnostic results, where Convolutional Neural Networks (CNNs), Transformers and Graph Neural Networks (GNNs) have been widely applied. In terms of automatic dental illness diagnosis, while CNN-based methods \cite{prajapati2017classification,lee2018detection,alalharith2020deep,tuzoff2019tooth} can effectively learn task-specific local image features for dental illness diagnosis, their capability in modelling global context information is limited, which lead to poor performances in dealing with complex tasks (e.g., diagnosing temporomandibular joint disorders or planning comprehensive orthodontic treatment) \cite{alzubaidi2021review} that require overall understanding. Consequently, transformer-based dental illness diagnosis approaches \cite{sheng2023transformer,almalki2023self} address this limitation by capturing global semantic interactions through self-attention \cite{shaw2018self}, but they require large-scale training data and computational resources. To mitigate small dataset issues, some Transformer methods use self-supervised learning (e.g., SimMIM and UM-MAE) \cite{almalki2023self} or multi-scale feature extraction [16]. Besides, the unique advantage of GNN-based approaches \cite{lian2019meshsnet,liu2023tucnet} lies in their ability to capture local and cues within the each dental object while facilitating interactions among all objects via edges for global information modelling \cite{wu2020comprehensive}, and thus have been widely explored for labeling the three-dimensional surfaces of teeth \cite{lian2019meshsnet,lian2020deep}. Although such  DL-based methods have been frequently applied to in dentistry such as tooth segmentation\cite{jader2018deep,sheng2023transformer}, tooth numbering\cite{tuzoff2019tooth}, detection of dental issues\cite{kim2019dentnet,muresan2020teeth} of instances of teeth and dental restorations, to the best of our knowledge, none of them has been developed for automated diagnosis of malocclusion issues.

Although the methods above show advantages in specific dental image applications, they require sufficient training examples for conditions like cavities and gum diseases, and may suffer from unstable training without large-scale data.
To facilitate developing effective deep learning models for automatic dental illness analysis, various types of dental image datasets have been collected and made publicly available, such as Digital Dental X-ray Database for Caries Screening \cite{rad2016digital}, Children's Dental Panoramic Radiographs Dataset\cite{zhang2023children}, DENTEX\cite{hamamci2023diffusion}, and DC1000 Dataset\cite{wang2023multi}. However, these datasets still suffer from various limitations for developing effective and robust DL-based dental analysis methods, including insufficient number of training examples \cite{rad2016digital,zhang2023children}, limited number of subjects involved in the data collection \cite{zhang2023children,rad2016digital}, unprofessional and non-standard labeling \cite{wang2023multi}, and single view of the recorded image \cite{rad2016digital,zhang2023children,hamamci2023diffusion,wang2023multi}. In addition, while some datasets can partially address the above issues, most of them rely on X-ray imaging, which is challenging due to radiation exposure and operational complexity, limiting scalability. More importantly, to the best of our knowledge, there is no dental image dataset specific collected for malocclusion, which further limits the development of DL models for automatic diagnosis of malocclusion issues.

In this paper, we present the first publicly available dental image dataset for the development of automated diagnostic models for malocclusion issues, the Oral and Maxillofacial Natural Image (OMNI) dataset. 
It contains 4166 images captured by an RGB camera from 384 patients that cover five different views, including frontal, left, right, maxillary and mandibular occlusal views, and annotated by professional dentists.
The OMNI dataset has largely solved the problem of shortage of medical data related to malocclusion issues, and provided comprehensive data support for research related to automatic localization and diagnosis of malocclusion issues. Our OMNI dataset stands out due to its data diversity, high volume of data, recorded from multiple views, and accurate annotations for malocclusion issues. In addition, we extend six standard deep learning models (including three CNNs, two transformers, and a GNNs-based method) to this dataset, providing the first study to evaluate the feasibility of applying deep learning to malocclusion issues diagnosis. The main contributions and novelties of this paper are summarized as follows:

\begin{enumerate}

\item We propose the first publicly available multi-view dental medical image dataset for malocclusion issues diagnosis (called OMNI Dataset). Our dataset opens up a new research avenue for multi-view dental image-based automatic malocclusion issues diagnosis.

\item  This papers conducted the first comprehensive analysis of and comparison between the performances of six widely-used CNNs, Transformers and GNNs for the task of automatic malocclusion issues recognition, showing that all of these DL models can accurately recognize various malocclusion issues to some extent.

\item This paper proposes a novel multi-dimensional edge GNN-based baseline (called GraphTeethNet) to dental image-based automatic malocclusion issues diagnosis, providing strong baselines for future researchers to develop their deep learning malocclusion issues diagnosis models.

\end{enumerate}

\section{Related Work}
\subsection{Existing dental image datasets}
There are already exists numerous dental image datasets \cite{silva2018automatic} for automated dental diagnosis and related tasks, which has driven the advancement of deep learning algorithms in dental diagnostics. 
However, most of these datasets only contain a small number of labelled images, and thus limited the development of deep learning models. For example, Children's Dental Panoramic Radiographs dataset \cite{zhang2023children} is the world's first pediatric dental panoramic radiograph dataset for caries segmentation and dental disease detection. It contains 100 high-quality panoramic radiographs (annotated) and 93 supplemental images (unannotated) collected from 106 pediatric patients. Digital Dental X-ray Database for Caries Screening \cite{rad2016digital} consists of 120 apical dental X-ray images, all digitized in grayscale format with a resolution of $748 \times 512$ pixels. 
Besides, there also exists several datasets including relatively large number of examples (i.e., more than a thousand labelled images). The DENTEX \cite{hamamci2023diffusion} includes three levels of panoramic radiographs: 693 radiographs labeled with quadrant information only, 634 radiographs labeled with quadrant and tooth number information, and 1005 radiographs labeled with quadrant, tooth number and diagnostic information (e.g., dental caries, deep caries, periapical lesions and blocked teeth). In addition, 1571 unlabeled radiographs were also provided for model pre-training. 
Finally, DC1000 dental caries dataset \cite{wang2023multi} contains 1000 panoramic radiographic images, of which 593 were accurately labeled while 407 were roughly labeled by five dentists, with a total of more than 7500 carious lesions covering three different levels of caries: shallow, moderate and deep caries. 

\subsection{Deep Learning  for Dental Image Analysis}
Recently, an increasing number of researchers are focusing on applying DL-based methods for automated diagnosis in the dental field. Currently, there are three main types of DL-based methods used in research on this area: CNN-based, Transformer-based and GNN-based. 

\begin{figure}
\centering
\includegraphics{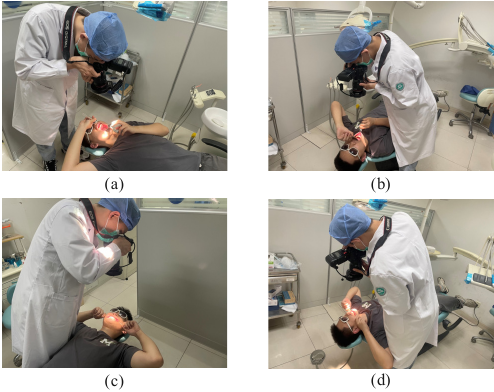}
\caption{Multi-view malocclusion dental image acquisition process. (a) the acquisition of maxillary maxillofacial photographs; (b) the acquisition of mandibular maxillofacial photographs; (c) the acquisition of frontal occlusal photographs; and (d) the acquisition of lateral occlusal photographs.}
\label{environment}
\end{figure}

\subsubsection{CNN-based methods} 
Various CNN-based methods have been proposed to dental image analysis, which can be roughly categorized as traditional CNN network-based, Faster R-CNN-based and Mask R-CNN-based. For example, Prajapat et al. \cite{prajapati2017classification} used transfer learning with a pre-trained VGG-16 \cite{simonyan2014very} model to classify dental diseases: dental caries, periapical infection, and periodontitis. Lee et al.\cite{lee2018detection} adapted GoogLeNet Inception v3 \cite{szegedy2016rethinking} using transfer learning for more accurate caries detection.
Alalharith et al. \cite{alalharith2020deep} proposed to automatically detect teeth and inflamed regions of the gums based on an improved Faster RCNN-ResNet50 feature extractor. 
Mahdi et al. \cite{mahdi2020optimization}, on the other hand, applied ResNet-50 and ResNet-101 as the backbones for the Faster R-CNN to achieve the automatic teeth detection and numbering. 
Alternatively, Anantharaman et al. \cite{anantharaman2018utilizing} proposed a Mask R-CNN \cite{he2017mask} based approach for the detection and segmentation of oral cold sores and canker sores. Jader et al. \cite{jader2018deep} applied Mask R-CNN for the first time on panoramic radiographic images for automatic detection and teeth instance segmentation.  Although CNN-based methods have made significant progress, they still share a major drawback: limited ability to handle complex deformations and multi-scale features. 

\subsubsection{Transformer-based methods}
Recently, transformer architecture also have been widely explored in dentistry to compensate for the shortcomings of CNNs. Both Jiang et al. \cite{jiang2021rdfnet} and Hossain et al. \cite{hossain2023cavit} proposed transformer-based methods for caries detection. 
Specifically, Jiang et al. \cite{jiang2021rdfnet} introduces the FReLU activation function to activate the visual-spatial information to improve the speed of caries detection. Hossain et al. \cite{hossain2023cavit} extended the Vision Transformer (ViT) \cite{dosovitskiy2020image} to process dental images recorded by smartphones, aiming to classify the images as advanced caries, early caries, or caries-free. 
Besides, Sheng et al. \cite{sheng2023transformer} proposed a U-shaped encoder-decoder transformer architecture with skip connections for teeth segmentation in panoramic radiographic images.  
Almalki et al. \cite{almalki2023self} applied the Swin-Transformer \cite{liu2021swin} for analyzing dental panoramic X-ray images. While the Swin-Transformer requires extensive data for training, and available dental X-ray images are very limited, they utilized self-supervised learning techniques to mitigate the issue of scarce data in panoramic X-ray images. Kadi et al. \cite{kadi2024detection} trained a Detection Transformer (DETR) \cite{carion2020end} to detect, segment teeth for 52 different dental categories, and proposed a stable diffusion-based data augmentation techniques. 

\subsubsection{GNN-based methods}
Compared to the previous CNN and transformer-based methods, GNN-based methods can flexibly handle complex tooth geometries, reduce information loss with high computational efficiency \cite{han2022vision}, which makes them suitable for applications such as tooth reconstruction and diagnosis.
Due to the difficulty of labeling the 3D surfaces of teeth, it is challenging to accommodate the variations in dental appearances across different patients.
Consequently, Lian et al. \cite{lian2019meshsnet} proposed a deep multi-scale mesh feature learning method, MeshSNet, for end-to-end 3D teeth surface annotation. It employs a novel graph-constrained learning module that hierarchically extracts multi-scale contextual features, and then densely integrates both local to global geometric features, which comprehensively characterizes the mesh elements for segmentation tasks. They further extend this approach as a MeshSegNet \cite{lian2020deep} which applies a Multilayer Perceptron (MLP) to extract the higher-order geometric features of the mesh cells. It aligns the initial features by means of a Feature Transformation Module (FTM), and progressively increases the receptive domain through the multi-scale Graph constraint Learning Nodule (GLM) to capture the multi-scale local geometric features in a hierarchical manner. 

\section{OMNI Dataset}
In this paper, we establish the first publicly available multi-view maxillofacial medical image dataset (called Oral and Malocclusion Natural Image (OMNI) Dataset) for the development of deep learning-based automatic malocclusion issues diagnosis systems. It includes 4166 RGB images of the oral cavity captured by a standard RGB camera, with an image size of $512 \times 341$, recorded from multiple views, including frontal occlusal, left occlusal, right occlusal, maxillary craniofacial, and mandibular craniofacial. All images are labeled with multiple malocclusion issues by professional dentists.

\subsection{Data collection protocol}
\label{subsec:hu}

\subsubsection{Participants and equipment setup}
Our OMNI dataset collection is conducted in the Department of Stomatology of the Third Affiliated Hospital of Soochow University (the First People's Hospital of Changzhou), under the supervision of professional doctors. Data were gathered from patients visiting for routine consultation and treatment, with images taken during the first visit and two follow-up visits to ensure data continuity and comprehensiveness. As shown in Fig. \ref{environment}, the equipment used in our data collection include a dental chair, a retractor, a dental light curing lamp and a Canon camera (model EOS 550D) for the image acquisition.

\subsubsection{Data collection scenario}
\begin{figure}
\centering
\includegraphics[width=9cm]{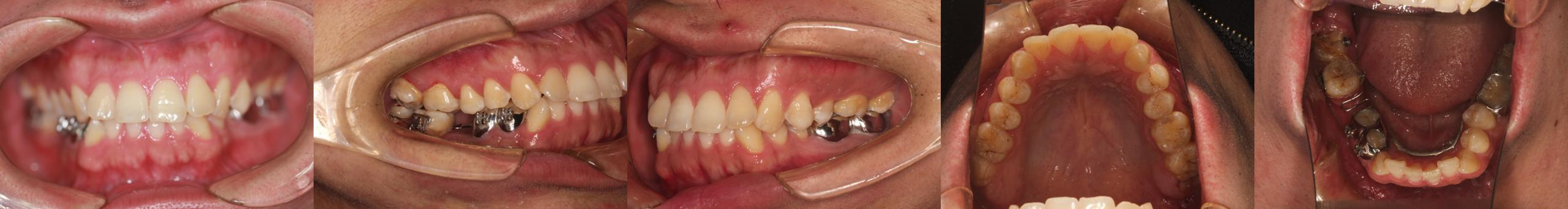}
\caption{Sample images of teeth from five different views. From left to right: frontal, right side, left side, maxillary and mandibular view.}
\label{Examples dental images}
\end{figure}

In our dataset, a total of five views of the images are acquired for each patient. Prior to the data collection, each patient was asked to perform dental hygiene and cleaning. For patients wearing a removable orthodontic appliance, they were asked to wear the appliance during the collection. Subsequently, the patient is asked to lie down in the dental chair while the doctor took photos for the patient's teeth using an EOS 550D Canon camera under a natural light condition, which is set to the M-stop, with the flash to 1/4-stop. Specifically, the photos were taken with the camera positioned perpendicular to the surface of the patient's teeth, allowing that the lens is aligned at a 90-degree angle to the teeth surfaces. For taking frontal photos, each patient was ask to lie flat on a dental chair, and then support the oral mucosa themselves by using a retractor and lip rest, allowing retraction of the lips and cheeks. This setup ensures  patients to expose the surface of their teeth (both upper and lower), while maintain an occluded bite. When taking left and right side occlusion photos, each patient was required to tilt his/her head to one side and force the mouth corners on the other side by a pulling hook. In the recording of maxillary and mandibular photographs, each patient is asked to keep the mouth open while a mirror was placed in the mouth to reflect the upper and lower dental arches, In order to clearly capture the entire dental arch from the top (upper arch) or bottom (lower arch). To acquire high-quality dental images, an angle puller was also used to ensure the teeth of each acquired image is centered and symmetrical. As a result, the acquiring processes for dental images of four different views focus on the following operations:
\begin{itemize}
    \item \textbf{Frontal photographs:} centering the dental midline and ensuring the symmetry along the occlusal plane, where the angle of the mouth hooks is kept in a straight line with the occlusal plane, and the upper lip should be centered on the midline to ensure the symmetry of the left and right buccal mucosa and the even distribution of the number of teeth. 

    \item \textbf{Left and right side photographs:} the centering is focused on the buccal surfaces of the teeth, where the centers of the hooks is flushed with the occlusal plane. Specifically, the hooks on the non-photographic side not touch the anterior teeth, while the hooks on the photographic side is spread as far as possible to ensure that the cuspids are centered. 

    \item \textbf{Maxillary photographs:} the images are centered on the dental arch and symmetrical along the dental midline, where a reflecto that does not touch the last molar is used, with the edge of the reflector aligned with the left and right molar gaps and the hooks spread wider than the reflector.

    \item \textbf{Mandibular photographs:} the images are centered on the dental arch and symmetrical along the dental midline, where the angle of the reflector is set to as wide as possible to avoid covering the dorsum of the tongue, as well as ensure that the occlusal surfaces of the first molars are in focus.
\end{itemize}
Fig. \ref{environment} visualizes our multi-view dental image acquisition scenarios. Example dental images recorded from these views are visualized in Fig. \ref{Examples dental images}.

\subsubsection{Relevant statistics}
Our OMNI dataset consists of 4166 RGB dental images from 384 Chinese participants (153 males, 231 females), aged 3-48 years, with a mean age of 10.4 years old and a standard deviation of 5.63. It includes 903 frontal occlusal, 841 left occlusal, 843 right occlusal, 820 maxillary maxillofacial, and 759 mandibular maxillofacial images. A total of 10 classes related to malocclusion have been labeled, where Table \ref{tb: number of categories} lists the name and the number of images suffered from each type of issue. Except of the healthy teeth (HT class), among all the malocclusion disorders, the most frequently appeared malocclusion issue is the Tooth Torsion (TT), while the Cast Fixed Orthodontic Appliances (CFOA) issue is less likely to be occurred. In our OMNI dataset, there is an average of 1.03 malocclusion issues appeared in each image, including 565 images with no teeth suffer from any malocclusion issues. Among the images suffering from malocclusion, a minimum of one issue and a maximum of three issues are presented in each of them, where 2945 images suffer only one issue; 603 images suffer from two issues, and 53 images contain three issues.

\begin{table}[t]
\caption{Malocclusion issue categories and their example numbers annotated in our OMNI dataset.}
\centering
\begin{tabular}{ccc}
\toprule
Classes & Specialized medical terms & Quantities\\
\midrule
HT & Healthy Teeth & 3610\\
TT & Tooth Torsion & 1686\\
DO & Deep Overjet & 205\\
IOA & Invisible Orthodontic Attachment & 402\\
TE & Tooth Emergence & 441\\
CFOA & Cast Fixed Orthodontic Appliances & 144\\
TM & Tooth Misalignment & 147\\
MR & Mandibular Retrusion & 289\\
OB & Orthodontic Brace & 220\\
FOD & Fixed Orthodontic Device & 776\\
\bottomrule
\end{tabular}
\label{tb: number of categories}
\end{table}

We divided our dataset into three sets, including training, validation, and test set, resulting in 2,481 training images (534 frontal, 497 left, 503 right, 492 maxillary, and 455 mandibular occlusal images), 857 validation images (187 frontal, 174 left, 174 right, 167 maxillary, and 155 mandibular occlusal images), and 828 test images (182 frontal, 170 left, 166 right, 161 maxillary, and 149 mandibular occlusal images), where the training set is used for learn models, the validation set aims for hyper-parameter tuning and model selection, while the test set is established for the final performance evaluation.

\subsubsection{Data privacy and ethical consideration}
Strictly adherence to ethical and privacy-preserving protocols is essential for the collection of medical datasets. To not only ensures the security and privacy of the patient's personal information, but also safeguards the legality and compliance of the data collection process, we strictly adhere to the guiding principles of the Declaration of Helsinki \cite{williams2008declaration}. Specifically, before the data collection, we ensured that patients who participating in our study are fully informed about the purpose, methodology, expected benefits, and possible risks of it. At the same time, patients whose dental images are being collected will be asked to voluntarily sign an informed consent form and choose whether or not to allow their dental images to be used and published for academic purposes.

\subsection{Annotation}
\label{subsec:hu}
The annotation for our OMNI dataset is achieved by three key steps: sample selection, teeth localization, and categorical malocclusion issues annotation. The sample selection filters out low-quality dental images to ensure the clarity and integrity of the data. Then, the teeth localization provides accurate tooth locations for each image. Finally, categorical malocclusion issues are labelled for each image, which has been further validated through a annotation refinement process. This structured annotation approach ensures the high standard and quality of our OMNI dataset.

\subsubsection{Sample selection and data pre-processing}
\label{subsec:dft}
For better experimental results, we pre-processed the collected data by reviewing each image for quality. Poor-quality images were removed based on two criteria: \textbf{blurred images caused by movement}, which hinder clear recognition of tooth structure, and \textbf{images with incomplete dental structures}, which including images contain missing or partially obscured teeth, as well as images captured with insufficient capture range or abnormal tooth positioning.

\subsubsection{Teeth localization}
\label{subsec:dft}
\begin{figure}
\centering
\includegraphics[width=9cm]{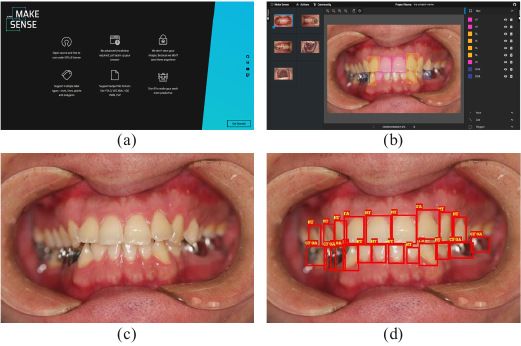}
\caption{The annotation tool and annotated examples. (a) Makesense.ai's operation page; (b) example of the annotation process; (c) the original dental image; and (d) the annotated dental image}
\label{annotation}
\end{figure}

We employed a powerful free online image annotation tool (called Makesense.ai \footnote{\url{https://www.makesense.ai/}}) to annotate every image of our OMNI dataset, including the position and category of each tooth. As shown in Fig. \ref{annotation}, the dentists use Makesense.ai to initially label the teeth in each image i.e., drawing a bounding box to locate each tooth, which fits the tooth contours.

\subsubsection{Categorical malocclusion issues annotation}

All malocclusion issue types are annotated according to the seventh edition of Orthodontics published by People’s Medical Publishing House \cite{zhao2020orthodontics}. The following categories were annotated:
\begin{itemize}
    \item (1) \textbf{HT:} Teeth are properly aligned and functioning within the dental arch, which exhibit no signs of decay, periodontal disease, or structural damage;

    \item (2) \textbf{TT:} A condition where a tooth is rotated or twisted around its longitudinal axis, resulting in an abnormal orientation within the dental arch;

    \item (3) \textbf{DO:} A dental malocclusion where the upper front teeth (maxillary incisors) horizontally overlap the lower front teeth (mandibular incisors) to an excessive degree;

    \item (4) \textbf{IOA:} These attachments, part of clear aligner or aesthetic orthodontic treatments, are tooth-colored or transparent devices bonded to teeth to aid in controlled tooth movement without the visibility of traditional braces.

    \item (5) \textbf{TE:} The process by which a tooth moves through the alveolar bone and soft tissue to appear in the oral cavity. This is a natural phase of dental development.

    \item (6) \textbf{CFOA:} Custom-made orthodontic devices, such as fixed space maintainers or lingual braces, which are fabricated using dental impressions (casts) of the patient's teeth and are permanently attached to the teeth to correct or maintain dental alignment;

    \item (7) \textbf{TM:} improper positioning of teeth within the dental arch, leading to malocclusion. This includes various forms of malpositions such as crowding, spacing, and deviations from the ideal arch form;

    \item (8) \textbf{MR:} A skeletal condition characterized by the posterior positioning of the mandible relative to the maxilla, often contributing to the class II malocclusion;

    \item (9) \textbf{OB:} A comprehensive orthodontic appliance consisting of brackets, wires, and sometimes bands, which is fixed to the teeth. This device applies continuous pressure to move teeth into the desired position;

    \item (10) \textbf{FOD:} Any orthodontic apparatus that is bonded to the teeth and remains in place throughout the treatment period. These devices, which include traditional braces and certain types of retainers, are used to correct dental and skeletal discrepancies by exerting controlled forces over time.
    
\end{itemize}

\subsubsection{Annotation refinement}
To ensure the consistency of annotations, uniform labeling standards have been applied to our annotation process, which are: (i) the bounding box should closely follow the tooth contour, minimizing the inclusion of redundant background areas as much as possible; and (ii) the status of the tooth should be accurately labeled based on the tooth’s cosmetic features and clinical information. Here, the initial annotation is performed by specialized doctors who are familiar with the annotation tools and standards. 

Specifically, after the initial annotation, a comprehensive and rigorous review mechanism is employed to ensure the quality of labels (i.e., location and malocclusion issues) for each image, which was performed by senior dentists and dental professionals. For \textbf{teeth locations}, a manual inspection conducted by experienced dentists carefully evaluates the size and position of the bounding boxes to ensure that they accurately cover the target tooth structure. Another team of senior dentists then review the annotations again to confirm that they are perfectly aligned with the target tooth structure. Then, \textbf{categorical malocclusion issue} annotations are assessed by a multi-criteria evaluation system defined by clinically validated guidelines, namely the American Board of Orthodontics (ABO) classification standards. In this process, dentists are required to categorize malocclusion issues according to these guidelines. The initial annotations are compared with a standardized checklist that includes criteria for the ten categories of issues in our OMNI dataset. If there are anomalies or discrepancies in these annotations, the corresponding annotations would be further reviewed by a panel of senior dental specialists and correct the annotations to ensure that they meet the established orthodontic diagnostic criteria. Furthermore, the annotation data were ultimately standardized in the COCO format, ensuring compatibility with mainstream deep learning frameworks and facilitating a streamlined model development workflow.

\section{Deep learning benchmark}
This section introduces six benchmark models for RGB image-based automatic malocclusion diagnosis, including three CNN-based, two Transformer-based, and one GNN-based model. These models establish the first deep learning baselines for automatic localization and assessment of malocclusion issues.

\subsection{CNN-based baselines}
\label{subsec:hu}
\textbf{Faster R-CNN.} Faster R-CNN \cite{ren2016faster} is an object detection model, 
which introduces a pre-trained CNN (e.g., ResNet) as a feature extractor, followed by the RPN generating candidate regions, which are then unified to a fixed size using ROI Pooling for classification and bounding box regression. In the context of automatic malocclusion issues diagnosis, Faster R-CNN leverages its high-precision object localization capability to accurately detect the position of each tooth, making it particularly suitable for small object detection tasks. 

\textbf{Mask R-CNN.} Mask R-CNN \cite{he2017mask} is an extension of Faster R-CNN that adds instance segmentation functionality, allowing it to generate pixel-level segmentation masks in addition to classification and bounding box regression. In malocclusion issues diagnosis, Mask R-CNN is used solely for its improved detection capabilities, without employing the instance segmentation component. 

\textbf{EfficientDet.} EfficientDet \cite{tan2020efficientdet} is an efficient object detection model that combines EfficientNet \cite{tan2019efficientnet} as the backbone network with a bidirectional feature pyramid network (BiFPN). The model utilizes multi-scale feature fusion to achieve high-efficiency object detection, making it particularly suitable for real-time detection and analysis of large-scale medical image datasets in malocclusion issues diagnosis. 

\subsection{Transformer-based baselines}
\label{subsec:hu}

\textbf{DETR.} DETR \cite{carion2020end} is an innovative object detection model based on a Transformer encoder-decoder architecture, designed to capture global contextual information in images. It treats object detection as a sequence-to-sequence prediction problem. In malocclusion issues diagnosis, DETR leverages its strong global feature modeling capabilities and adaptability to complex scenarios to effectively identify intricate tooth arrangements and subtle features, making it suitable for recognizing and diagnosing different types of malocclusions. 

\textbf{Deformable DETR.} Deformable DETR \cite{zhu2020deformable} is an improved version of DETR, introducing a deformable attention mechanism that allows the model to selectively focus on features at different scales and locations on the feature map, thereby enhancing the model's efficiency and accuracy. In malocclusion issues diagnosis, Deformable DETR enhances convergence speed and small object detection capabilities, making it particularly effective in handling the varying positions and shapes of teeth.

\subsection{Graph-based baseline}
\label{subsec:spe}
\begin{figure*}
	\centering
	\includegraphics[width=18.6cm]{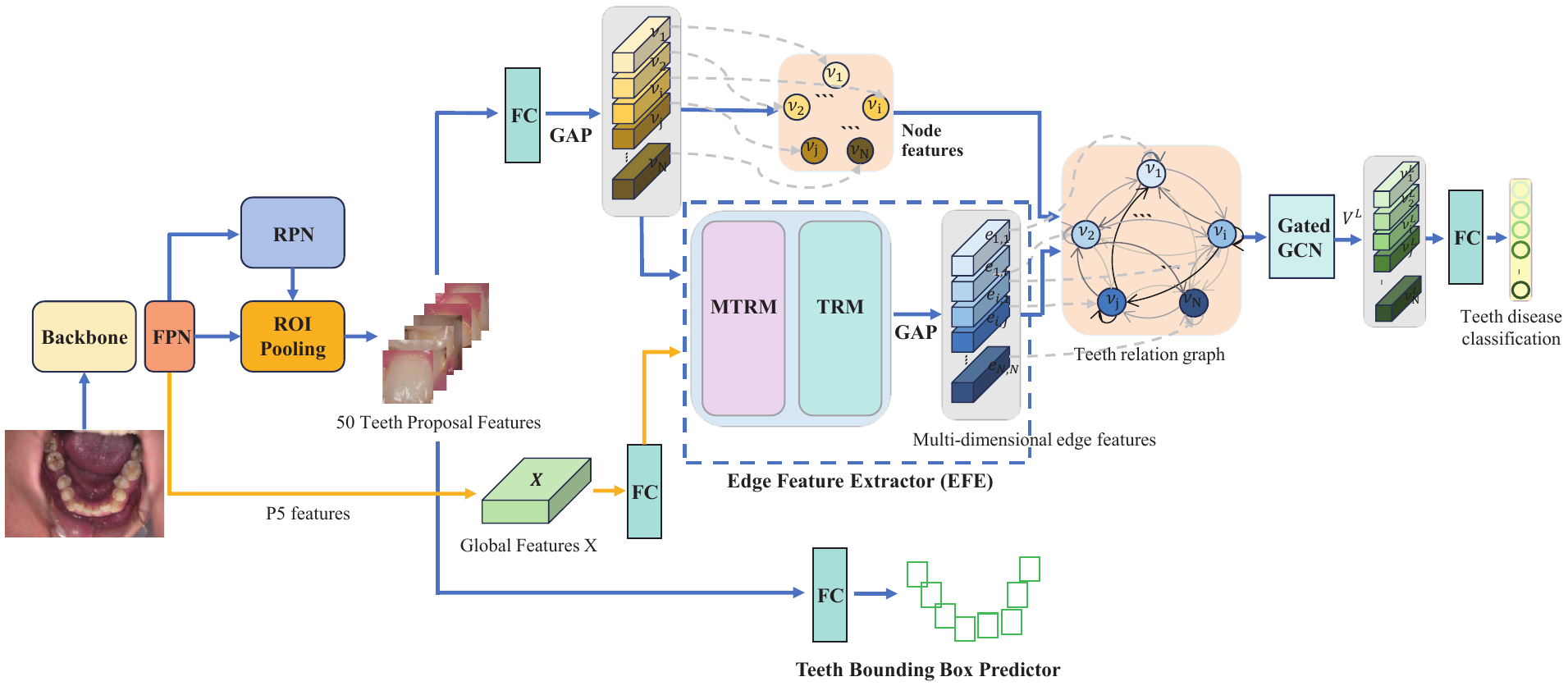}
	\caption{Illustration of our GraphTeethNet baseline.}
	\label{model}
\end{figure*}

\textbf{GraphTeethNet.} In addition to CNN and Transformer-based baselines, we additionally propose a graph-based baseline called GraphTeethNet (shown in Fig. \ref{model}), which is inspired by the generic GRATIS \cite{song2022gratis,luo2022learning}. Our method is based on Faster R-CNN. From ROI Pooling, we obtain $T_{box}$ teeth proposal features (restricted to 50 to facilitate subsequent computation of multi-dimensional edge features). The lower branch functions as a Teeth Bounding Box Predictor, responsible for predicting the coordinates of the teeth bounding boxes. The upper branch focuses on teeth disease classification, where the 50 teeth proposal features are treated as node features $\bm{V}_{i}$. These are combined with global features extracted from the P5 level of the FPN. Using the Edge Feature Extractor (EFE) module further learns global context-aware multi-dimensional edge features to describe the relationship between each pair of related teeth.

\textbf{Edge Feature Extractor:} After obtaining teeth node features, the EFE module of our GraphTeethNet learns a multi-dimensional edge feature to describe each edge, which aims to explicitly represent multiple task-specific relationship clues between different node features. It consists of two main blocks: Maxillofacial Teeth Representation Modelling (MTRM) and Teeth Relationship Modelling (TRM).
\begin{itemize}
	\item \textbf{MTRM.} The MTRM block takes the specific features of a pair of teeth node features, $\bm{V}_{i}$ and $\bm{V}_{j}$, as well as the global representation of the original dental image, $\bm{X}$, as the input. It first performs cross-attention between $\bm{V}_{i}$ and $\bm{X}$ as well as $\bm{V}_{j}$ and $\bm{X}$ separately, where the tooth-specific feature mappings $\bm{V}_{i}$ and $\bm{V}_{j}$ are used solely as queries, and the representation $\bm{X}$ of all teeth in the entire oral cavity is used as both key and value. Firstly, cross-attention is defined as: 
	\begin{equation}
		\text{MTRM}(\bm{Y}, \bm{Z}) = \text{softmax}\left( \frac{\bm{Y} \bm{W}_q (\bm{Z} \bm{W}_k)^T}{\sqrt{d_k}} \right) \bm{Z} \bm{W}_v,
	\end{equation}
	Thus, the whole process is formulated as follows: 
	\begin{equation}
		\bm{\mathcal{F}}^{TS}_{i,x}, \bm{\mathcal{F}}^{TS}_{j,x} = \text{MTRM}(\bm{V}_i , \bm{X}), \text{MTRM}(\bm{V}_j, \bm{X}),
	\end{equation}
	where $\bm{W}_q$, $\bm{W}_k$ and $\bm{W}_v$ are learnable weights and $\mathit{d}_{\mathit{k}}$ is a scaling factor equivalent to the channels of the "key".
	
	\item \textbf{TRM.} After passing through the MTRM block, each tooth has been independently encoded with specific cues. Therefore, the TRM is needed to further extract the relevant cues between the teeth. The cross-attention is used again between $\bm{\mathcal{F}}^{TS}_{i,x}$ and $\bm{\mathcal{F}}^{TS}_{j,x}$ to generate $\bm{\mathcal{F}}^{TR}_{i,j,x}$ and $\bm{\mathcal{F}}^{TR}_{j,i,x}$. $\bm{\mathcal{F}}^{TR}_{i,j,x}$ is generated using $\bm{\mathcal{F}}^{TS}_{j,x}$ as queries, and $\bm{\mathcal{F}}^{TS}_{i,x}$ as key and value, while $\bm{\mathcal{F}}^{TR}_{j,i,x}$ is generated in the opposite manner. Thus, $\bm{\mathcal{F}}^{TR}_{i,j,x}$ summarize the relevant cues for the teeth in $\bm{\mathcal{F}}^{TS}_{i,x}$, and $\bm{\mathcal{F}}^{TR}_{j,i,x}$ summarize the relevant cues for the teeth in $\bm{\mathcal{F}}^{TS}_{j,x}$. Finally, the obtained $\bm{\mathcal{F}}^{TR}_{i,j,x}$, $\bm{\mathcal{F}}^{TR}_{j,i,x}$ are input into GAP layer to obtain the multi-dimensional edge feature vectors $\mathit{e}_{\mathit{i,j}}$ and $\mathit{e}_{\mathit{j,i}}$. In this way, the edge feature vectors $\mathit{e}_{\mathit{i,j}}$ and $\mathit{e}_{\mathit{j,i}}$ summarize multiple dental cues from the whole maxillofacial image. The above process can be summarized as the following mathematical process: 
	\begin{equation}
		\mathit{e}_{\mathit{i,j}},\mathit{e}_{\mathit{j,i}} = \text{GAP}(\text{TRM}(\bm{\mathcal{F}}^{TS}_{j,x}, \bm{\mathcal{F}}^{TS}_{i,x}),\text{TRM}(\bm{\mathcal{F}}^{TS}_{i,x}, \bm{\mathcal{F}}^{TS}_{j,x}))
	\end{equation}
	
\end{itemize}
After the above two blocks, the teeth relationship graph $\bm{G}^{0} = (\bm{V}^{0},\bm{E}^{0})$ is learned, where $\bm{G}^{0}$ contains $\mathit{N}$ node features and $\mathit{N} \times \mathit{N}$ multi-dimensional directed edge features comprehensively describing the task-specific relationships between each pair of teeth.

\subsection{Evaluation Metrics}
To evaluate the performance of the trained model, we used mean Average Precision (mAP), following standard object detection practices.

\begin{itemize}
	
	\item \noindent \textbf{Mean Average Precision (mAP).} The mAP evaluates the overall performance of the model to detect all categories by calculating the average precision (AP) for each category under multiple IoU thresholds. For each category, its AP is obtained by calculating the area under the precision-recall curve for that category, which can be formulated as:
	\begin{equation}
		\text{AP} = \sum_n (R_n - R_{n-1}) P_n,
	\end{equation}
	where $P_n$ and $R_n$ are the precision and recall at the $\mathit{n}_{\mathit{th}}$ threshold, respectively. To calculate the mAP, we average the APs of all categories, which can be formulated as:
	\begin{equation}
		\text{mAP} = \frac{1}{C} \sum_{i=1}^C \text{AP}_i,
	\end{equation}
	where $\mathit{C}$ is the total number of categories. 
	
\end{itemize}

\section{Experiments}
\subsection{Data pre-processing}
\label{sub:dataset}
For all baselines, we performed a standard data pre-processing to ensure the fair comparison of them. Specifically, it consists of three main steps: 
(1) First, the original images are resized to $1333\times800$ with the aspect ratio preserved for Faster R-CNN, Mask R-CNN, DETR, Deformable DETR and GraphTeethNet. In contrast, for EfficientDet, input images are resized to a fixed resolution (e.g., $512\times512$ for EfficientNet-B0, $896\times896$ for EfficientNet-B3) without preserving the original aspect ratio. This is required to ensure that the feature maps at different scales can be properly aligned within the BiFPN structure.
(2) Then, a random horizontal flip with a probability of 0.5 is applied during training to augment the data and improve the model’s robustness to mirrored inputs.
(3) Finally, image normalization is performed using the mean $\mu$ and standard deviation $\sigma$ values from the ImageNet dataset \cite{deng2009imagenet}, to ensure compatibility with the pre-trained backbone network.
\begin{equation}
	I_{\text{normalized}} = \frac{I - \mu}{\sigma},
\end{equation}
where $\mathit{I}$ is the original image.

\subsection{Implementation details}
In our experiments, the GraphTeethNet model was trained and evaluated using the PyTorch framework, while the other models were trained and evaluated using the MMDetection \cite{mmdetection} framework. Specifically, Faster R-CNN, Mask R-CNN, EfficientDet, and Deformable DETR were all trained for 50 epochs, while DETR was trained for 300 epochs, and GraphTeethNet was trained for 100 epochs. The aforementioned models all used the AdamW optimizer with an initial learning rate of 1e-4 and a weight decay of 0.05. The learning rate scheduling consisted of an initial LinearLR warm-up phase, followed by a MultiStepLR scheduler with predefined milestones. The loss functions for each model are summarized in Table \ref{tb:loss}.
These varied configurations allowed us to comprehensively assess the performance and convergence of each model. The training process was evaluated on the validation set, and the model with the best performance was selected for testing on the test set. The corresponding epoch number of the best-performing model is provided in parentheses in Table \ref{tab:performance_comparison}. All experiments were conducted on NVIDIA GeForce RTX 4090.

\begin{table}[h]
	\caption{Loss Functions for Different Models}
	\centering
	\begin{tabular}{ccc}
		\toprule
		Methods & Classification Loss & Bounding box regression Loss \\
		\midrule
		Faster R-CNN & Cross-entropy & L1 \\
		Mask R-CNN & Cross-entropy & L1 \\
		EfficientDet & Focal & Smooth L1 \\
		DETR & Cross-entropy & L1 and Generalized IoU \\
		Deformable DETR & Focal & L1 and Generalized IoU \\
		GraphTeethNet & Cross-entropy & Smooth L1 \\
		\bottomrule
	\end{tabular}
	
	\label{tb:loss}
\end{table}

\subsection{Baseline results}
\begin{figure}
	\centering
	\includegraphics{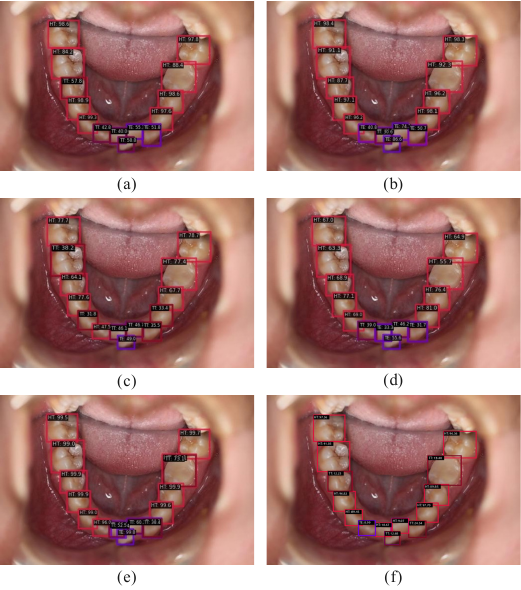}
	\caption{Example diagnosis results for all baseline models achieved for our OMNI dataset: (a) Faster R-CNN; (b) Mask R-CNN; (c) EfficientDet; (d) Deformable DETR; (e) DETR; (f) GraphTeethNet.}
	\label{prediction}
\end{figure}

This section conducts comprehensive experiments to evaluate the feasibility and performance of applying our baseline deep learning models to the task of automatic malocclusion issue diagnosis on our OMNI dataset. Table \ref{tab:performance_comparison} show that Deformable DETR achieves the highest mAP@0.5 across models, while Mask R-CNN's performance is a close second. 
The automated diagnostic results for each model are shown in Fig. \ref{prediction}.

\textbf{Category-wise mAP Analysis.} Table \ref{tab:performance_comparison} provide a detailed comparison of mAP@0.5 for each category across all baseline models, and the validity of the models varied across categories. We assume this is caused by the varying complexities and challenges associated with each category. In particular, Deformable DETR consistently outperformed others in most categories. It achieved 93.64 in MR, 93.60 in FOD, and 89.94 in HT. Meanwhile, EfficientDet performs particularly well on DO, and GraphTeethNet on TM.

\newcolumntype{M}[1]{>{\centering\arraybackslash}m{#1}}
\begin{table*}[h]
	\centering
	\caption{Comparison of experimental results for the OMNI dataset on baseline models (mAP with IoU of 0.5: 0.95).}
	\begin{tabular}{ M{2.2cm} M{1.0cm} M{0.4cm} M{1.1cm} M{1.1cm} M{0.6cm} M{0.6cm} M{0.6cm} M{0.6cm} M{0.6cm} M{0.6cm} M{0.6cm} M{0.6cm} M{0.6cm} M{0.6cm} }
			\toprule
			Methods & Epoches & mAP & mAP@0.5 & mAP@0.75 & HT & TT & DO & IOA & TE & CFOA & TM & MR & OB & FOD \\
			\midrule
			Faster R-CNN & 50(16) & 37.56 & 65.32 & 39.91 & 86.89 & \textbf{41.35} & 48.46 & 58.25 & 31.78 & 82.43 & \underline{48.08} & 92.10 & 73.84 & 90.01 \\
			Mask R-CNN & 50(11) & 36.45 & \underline{65.88} & 36.42 & \underline{89.10} & 39.34 & 47.87 & 60.85 & 35.68 & \underline{84.64} & 46.39 & \underline{92.46} & 73.41 & 89.02 \\
			EfficientDet & 50(17) & \underline{37.67} & 64.16 & \textbf{41.13} & 88.75 & 39.17 & \textbf{57.33} & 60.45 & \underline{40.88} & 81.81 & 8.07 & 90.25 & \underline{82.40} & \underline{92.50} \\
			DETR & 300(117) & 36.57 & 63.39 & 39.20 & 84.41 & 36.85 & 42.83 & \underline{61.40} & \textbf{42.78} & 83.03 & 30.17 & 86.05 & 76.76 & 89.63  \\
			Deformable DETR & 50(19) & \textbf{38.26} & \textbf{66.39} & \underline{40.62} & \textbf{89.94} & 39.58 & \underline{54.23} & \textbf{66.21} & 34.85 & \textbf{85.69} & 23.56 & \textbf{93.64} & \textbf{82.65} & \textbf{93.60}  \\
			GraphTeethNet & 100(17) & 37.20 & 63.89 & 39.48 & 84.51 & \underline{39.77} & 49.51 & 54.34 & 35.61 & 75.65 & \textbf{51.81} & 87.74 & 72.39 & 87.55 \\
			\bottomrule
	\end{tabular}
	\label{tab:performance_comparison}
\end{table*}

\textbf{mAP under different IoUs.} Table \ref{tab:performance_comparison} detail the mAP, mAP@0.5, and mAP@0.75 for each model, providing a comprehensive comparison of performance across different IoU thresholds.
It can be observed that Deformable DETR achieves the best overall performance, demonstrating its stability and superiority in handling varying object scales and complex spatial layouts. 
EfficientDet follows closely in terms of mAP of 37.67 and achieves the highest mAP@0.75 of 41.13 among all methods, showing that its lightweight architecture based on EfficientNet excels in capturing fine-grained spatial features, particularly under high-precision evaluation. 
Faster R-CNN and Mask R-CNN exhibit comparable performance, whereas Mask R-CNN performs noticeably worse at higher IoU thresholds, with a lower mAP@0.75 of 36.42, suggesting challenges in convergence and fine-grained localization. 
The proposed GraphTeethNet model achieves a competitive mAP of 37.20, mAP@0.5 of 63.89, and mAP@0.75 of 39.48, outperforming DETR and closely approaching Faster R-CNN. This demonstrates the potential of modeling inter-tooth relations via graph-based learning, particularly in capturing structured domain knowledge.

\subsection{Ablation studies}
This section further analyze the contributions of some key modules/components of our baseline models and their impacts on the corresponding models for automatic teeth localization and malocclusion issues diagnosis on our OMNI dataset.

\subsubsection{RoI Pooling VS. RoI Align}
Table \ref{tb:RoI} compares the performance of RoI Pooling and RoI Align in Faster R-CNN and Mask R-CNN for diagnosing malocclusion. RoI Align outperforms RoI Pooling in all metrics (mAP, mAP@0.5, and mAP@0.75), improving localization and classification by providing more accurate feature alignment. For example, in Faster R-CNN, RoI Align achieves an mAP of 37.56, while RoI Pooling achieves 36.40. While RoI Pooling is computationally more efficient, it fails to capture fine details and edge complexities, making RoI Align more suitable for accurate malocclusion issues diagnosis.

\begin{table}[h]
	\centering
	\caption{The Impact of RoI Pooling and RoI Align on Faster R-CNN and Mask R-CNN.}
	\begin{tabular}{cccc}
			\toprule
			Methods & mAP & mAP@0.5 & mAP@0.75\\
			\midrule
			Faster R-CNN+RoI Pooling & 36.40 & 64.86 & \underline{37.58} \\
			Mask R-CNN+RoI Pooling & 35.84 & 64.43 & 35.83\\
			Faster R-CNN+RoI Align & \textbf{37.56} & \underline{65.32} & \textbf{39.91}\\
			Mask R-CNN+RoI Align & \underline{36.45} & \textbf{65.88} & 36.42\\
			\bottomrule
	\end{tabular}
	\label{tb:RoI}
\end{table}

\subsubsection{The role of multi-dimensional edge features for the graph baseline}
Table \ref{tb:edge features} demonstrates the importance of multi-dimensional edge features in GNNs, particularly for malocclusion issues in the GraphTeethNet model. Including multi-dimensional edge features improves performance across all metrics: mAP increases from 35.69 to 37.20, mAP@0.5 from 62.09 to 63.89, and mAP@0.75 from 38.96 to 39.48. These results highlight the critical role of multi-dimensional edge features in accurately locating and classifying malocclusions, enhancing the ability of the model to capture complex relationships between teeth, particularly at higher confidence thresholds. Edge features significantly boost classification accuracy, underscoring their value in complex diagnostic tasks.

\begin{table}[h]
	\centering
	\caption{The Role of Edge Features in GNNs.}
	\begin{tabular}{cccc}
			\toprule
			Methods & mAP & mAP@0.5 & mAP@0.75\\
			\midrule
			GraphTeethNet & \textbf{37.20} & \textbf{63.89} & \textbf{39.48}\\
			GraphTeethNet without edge features & 35.69 & 62.09 & 38.96\\
			\bottomrule
	\end{tabular}
	\label{tb:edge features}
\end{table}

\subsubsection{The Impact of Different Backbones on Models' Performances}

Table \ref{tb:different backbone} compares the performance of all models using different backbone networks. The Deformable DETR model with ResNet50 performs best, achieving the highest mAP@0.5 of 66.39. EfficientDet with EfficientNet-B3 performs well under mAP@0.75 (41.13), demonstrating strong detection accuracy at higher IoU thresholds. However, using deeper backbones for models like DETR, Deformable DETR, and GraphTeethNet results in performance degradation, with DETR experiencing the most significant drop, possibly due to ineffective utilization of the deeper network's potential. These results highlight the importance of selecting the right model and configuration based on task requirements and operational conditions to optimize performance and resource usage.

\newcolumntype{M}[1]{>{\centering\arraybackslash}m{#1}}  
\begin{table}[h]
	\centering
	\caption{Impact of different Backbone on each model.}
	\begin{tabular}{M{2.2cm}M{1.8cm}M{0.4cm}M{1.1cm}M{1.2cm}}
		\toprule
		Methods & Backbone & mAP & mAP@0.5 & mAP@0.75\\
		\midrule
		Faster R-CNN & ResNet50 & 36.73 & 64.19 & 38.68\\
		Faster R-CNN & ResNet101 & 37.56 & 65.32 & 39.91\\
		Mask R-CNN & ResNet50 & 36.45 & \underline{65.88} & 36.42\\
		Mask R-CNN & ResNet101 & 37.00 & 65.48 & 39.08\\
		EfficientDet & EfficientNet-B0 & 34.90 & 63.45 & 36.13\\
		EfficientDet & EfficientNet-B3 & \underline{37.67} & 64.16 & \textbf{41.13}\\
		DETR & ResNet50 & 36.57 & 63.39 & 39.20\\
		DETR & ResNet101 & 13.42 & 36.29 & 6.17\\
		Deformable DETR & ResNet50 & \textbf{38.26} & \textbf{66.39} & \underline{40.62}\\
		Deformable DETR & ResNet101 & 35.05 & 62.07 & 35.87\\
		GraphTeethNet & ResNet34 & 37.20 & 63.89 & 39.48\\
		GraphTeethNet & ResNet50 & 35.95 & 61.42 & 38.58\\
		\bottomrule
	\end{tabular}
	\label{tb:different backbone}
\end{table}

\section{Conclusion}
This paper introduces the OMNI dataset, the first oral and maxillofacial dataset designed for deep learning-based malocclusion diagnosis. It includes 4,166 images from five views and annotations for ten malocclusion issue categories by experienced dentists. The study benchmarks six deep learning models for teeth localization and malocclusion issue classification. 
The class imbalance in the dataset can affect model performance, and future work will focus on expanding the dataset, addressing class imbalance, and integrating multimodal data like 3D scans and X-rays for more robust diagnostics. In conclusion, the OMNI dataset and benchmarking provide a valuable foundation for future research and the development of advanced dental diagnostic models.

\bibliographystyle{IEEEtran}
\bibliography{egbib}

\end{document}